\pdfoutput=1

\documentclass[11pt]{article}

\usepackage{acl}

\usepackage{times}
\usepackage{latexsym}

\usepackage[T1]{fontenc}

\usepackage[utf8]{inputenc}

\usepackage{microtype}

\usepackage{inconsolata}

\usepackage{booktabs}
\usepackage{graphicx}
\usepackage{multirow}
\usepackage{enumitem}
\usepackage{longtable}

\title{Creating Suspenseful Stories: Iterative Planning with Large Language Models}

\author{Kaige Xie \\
  School of Interactive Computing \\
  Georgia Institute of Technology \\
  \texttt{kaigexie@gatech.edu} \\\And
  Mark Riedl \\
  School of Interactive Computing \\
  Georgia Institute of Technology \\
  \texttt{riedl@cc.gatech.edu} \\}

\begin{document}
\maketitle
\begin{abstract}
Automated story generation has been one of the long-standing challenges in NLP.
Among all dimensions of stories, {\em suspense} is very common in human-written stories but relatively under-explored in AI-generated stories.
While recent advances in large language models (LLMs) have greatly promoted language generation in general, state-of-the-art LLMs are still unreliable when it comes to suspenseful story generation.
We propose a novel iterative-prompting-based planning method that is grounded in two theoretical foundations of story suspense from cognitive psychology and narratology.
This theory-grounded method works in a fully zero-shot manner and does not rely on any supervised story corpora.
To the best of our knowledge, this paper is the first attempt at suspenseful story generation with LLMs.
Extensive human evaluations of the generated suspenseful stories demonstrate the effectiveness of our method.
\end{abstract}

\section{Introduction}

\begin{figure}[t]
\centering
\includegraphics[width=\linewidth]{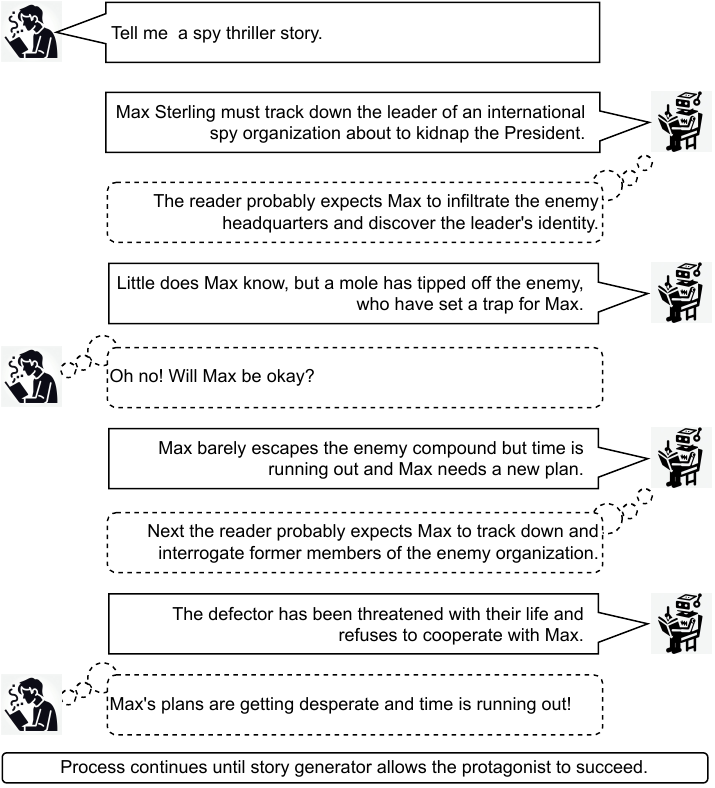}
\caption{Our suspenseful story generation technique is predicated on a theory of suspense in which the quantity and quality (likelihood) of ways in which a protagonist can avoid an undesirable outcome are decreased.
Our method iteratively produces possible actions for the protagonist to take and adversarially creates the conditions under which the protagonist will fail. The dialogue boxes indicate the story snippets generated by our method. The thought bubbles of the reader represent the mental process. The thought bubbles of the system depict the planning process in our method.
}
\label{fig:intro}
\end{figure}

Among numerous NLP tasks, automated story generation is a representative one that requires creativity.
By learning from human-written stories, an automated storyteller mimics humans and becomes competent in producing stories useful for various application scenarios, such as entertainment, education, and social bonding~\cite{riedl2010narrative}.
The notable achievements in the field of deep learning have led to the subsequent emergence of data-driven methodologies for automated story generation~\cite{martin2018event,clark-etal-2018-neural,fan-etal-2018-hierarchical,fan-etal-2019-strategies,yao2019plan}.
With the rapid development of large language models~\cite{radford2019language,brown2020language,openai2022chatgpt}, generated stories have further increased greatly in length, complexity, and fluency.
These enhancements are primarily realized through the application of methodologies that are built based on either fine-tuning pre-trained models on supervised story datasets or prompting large language models (LLMs) of strong zero-shot capability.

Existing work on language model based computational storytelling mostly focuses on optimizing automated story generation systems from different angles, including both generation settings (e.g., long-form generation~\cite{goldfarb-tarrant-etal-2020-content,yang-etal-2022-re3,yang2022doc}, controllable generation~\cite{peng-etal-2018-towards,lin2021plug,peng-etal-2022-guiding}), and story characteristics (e.g., commonsense reasoning~\cite{guan2019story,guan-etal-2020-knowledge,ammanabrolu2021automated}, temporal and causal relationships~\cite{guan-etal-2020-knowledge,ammanabrolu2021automated,han-etal-2022-go}).

In this paper, aiming to expand the capability of automated storytellers, we direct our focus towards an under-explored dimension of stories: \textit{suspense}, which is one of the strong affective responses humans often feel when being told a story.
Suspense is important as expert storytellers frequently use it to maintain reader engagement.
Due to the intricate nature of suspense as a complex cognitive phenomenon that only emerges under the right conditions, the generation of narratives with suspense, herein referred to as \textit{suspenseful story generation}, remains an open research challenge.
Prior research pertaining to narrative suspense~\cite{o2011toward,o2014dramatis,doust-piwek-2017-model,wilmot-keller-2020-modelling,zehe2023towards} predominantly focuses on the computational modeling of this cognitive phenomenon, with a very limited exploration of story suspense from the language generation perspective~\cite{cheong2014suspenser,fendt2016leveraging,delatorre2020predicting,delatorre2021improving}.
General-purpose generative models such as ChatGPT~\cite{openai2022chatgpt} exhibit a poor understanding of story suspense.
Model fine-tuning on suspenseful story corpora does not result in an LLM that is able to capture the latent conditions necessary for the reader to feel suspense.

Given the current widespread popularity of LLMs, how can one employ them to effectively generate suspenseful stories?
A psychological theory on suspense~\cite{gerrig1994readers} suggests that readers feel suspense when they believe that the protagonist is facing a negative outcome and that the quantity or quality of feasible means of escaping the impending negative outcome has become diminished.
Following this theory, we propose an \textit{iterative-prompting-based planning} technique that can effectively guide LLMs in suspenseful story generation.
Specifically, we establish a character with a goal and a negative outcome if they fail.
We prompt the LLM to consider the different ways the protagonist might achieve their goal.
We then adversarially attack these plans by establishing conditions in the story world in which the protagonist's plans will fail.
Repeating this several times produces an outline for the story, which we then elaborate into longer text sequences (chapters).
We depict and illustrate the overall intuition of our method in \autoref{fig:intro}.

In summary, the main contributions of our work are as follows:
\begin{itemize}
    \item We bring together the traditional narrative \& psychological theories on story suspense and the modern LLM-based language generation techniques. Our work is the first attempt at suspenseful story generation with LLMs.
    \item We propose an iterative-prompting-based planning technique for suspenseful story generation. This theory-grounded technique works in a fully zero-shot manner and does not rely on any supervised story corpora.
    We perform extensive human evaluations and in-depth analyses that demonstrate the effectiveness of our method in automatically creating suspenseful stories with the LLM.
    \item Using our validated method, we further advance the understanding of the factors that contribute to reader perceptions of suspense through controlled intervention studies.
\end{itemize}

\section{Related Work}

\subsection{Automated Story Generation}

Early work on story generation used symbolic planning~\cite{meehan1976metanovel,lebowitz1987planning,cavazza2003interacting,porteous2009controlling,riedl2010narrative,ware2010modeling,ware2021sabre}.
These systems require substantial knowledge engineering of logical constraints, limiting their generality, and don’t always generate plots or stories in natural language.

Neural language modeling approaches~\cite{roemmele2016writing,khalifa2017deeptingle,martin2018event,clark-etal-2018-neural,yao2019plan,fan-etal-2019-strategies,rashkin-etal-2020-plotmachines,ammanabrolu2021automated,jin-etal-2022-plot} have been applied to story generation because they circumvent the need for manual knowledge engineering and tend to produce relatively fluent, varied, and naturalistic language.
A significant amount of work focuses on the controllability of generators~\cite{peng-etal-2018-towards,lin2021plug,peng-etal-2022-guiding}.
Others focus on improving the system from different dimensions of story goodness, e.g., commonsense reasoning~\cite{guan2019story,guan-etal-2020-knowledge,ammanabrolu2021automated} and temporal and causal relationships~\cite{guan-etal-2020-knowledge,ammanabrolu2021automated,han-etal-2022-go}.

Large, pre-trained language models such as GPT-3, ChatGPT, and GPT-4, are capable of generating longer, more fluent story sequences.
Generation can be extended to many thousands of words~\cite{yang-etal-2022-re3,yang2022doc}.
However, LLMs have been unreliable when it comes to generating novel, suspenseful stories.
One reason for this is that suspense is a complex cognitive phenomenon that does not emerge in the latent state representations of a transformer.
Instead, one must infer the cognitive state of the reader, or induce a narrative structure with known properties that co-occur with suspense.

There exist works on suspenseful story generation~\cite{cheong2014suspenser,fendt2016leveraging,delatorre2020predicting,delatorre2021improving}, but they all require substantial knowledge engineering, making them limited in generality.

\subsection{Story Suspense}

Suspense is a complex affective response to the events in a narrative that combines hope simultaneously with fear coupled with a cognitive state of uncertainty~\cite{ortony_clore_collins_1988}.
Why do some stories evoke feelings of suspense when others do not?
~\citet{gerrig1994readers} provide an accounting of suspense backed up by psychological studies.
First, the reader must be {\em transported} into the fictional world. 
{\em Transportation} refers to the phenomenon of feeling immersed in the story world~\cite{gerrig1993experiencing}.
Transportation is accompanied by a dampening of critical reasoning about facts other than those presented in the story (for example, we know that James Bond will not be killed, but we become anxious about his prospects of surviving a mission anyway).
Second, the reader is given a reason to hope for an important target outcome.
This could be a very positive state for the protagonist (the protagonist gets together with their love interest), or the avoidance of a very negative state (the hero is not killed).
Typically, the reader develops empathy for the protagonist~\cite{jose1984development}.
Third, the reader engages in problem-solving to figure out how the protagonist can achieve their goal (or avoid the negative state).
Fourth, the space of actions the protagonist can take must have few paths to success, and/or those paths must be seen as having a low probability of success.

Another factor related to suspense is the narratological theory of {\em disparities of knowledge}~\cite{branigan2013narrative}.
This framework proposes that readers and story characters can have different knowledge of the story world.
When a character knows something that the reader doesn't know, the revelation of that information can invoke surprise in the reader.
However, when the reader knows something that the story world characters don't know, and that knowledge is related to the protagonist's chances of a negative outcome, then the reader may feel suspense.
Thus, a key consideration in suspense generation is reasoning about what characters know versus what readers know.

\citet{o2014dramatis} attempt to create a computational model of the accounting of suspense~\cite{gerrig1994readers}.
In order to determine whether a story might be perceived as suspenseful, they generate plans for the protagonist and analyze the quantity and quality of possible plans.
However, since they use a symbolic planner, they must know the characters and possible actions in advance, limiting practical applicability.
Their detection technique does not lend itself to story generation.
\citet{doust-piwek-2017-model} propose a graph analysis approach to suspense detection.
\citet{wilmot-keller-2020-modelling} propose a hierarchical language-modeling approach to suspense detection.
\citet{zehe2023towards} propose to detect suspense by training on a corpus of dangerous situations.

\section{Iterative-Prompting-based Planning}
\label{sec:method}

In this section, we introduce our method for generating suspenseful stories with an LLM: \textit{Iterative-Prompting-based Planning}.
One of the reasons that LLMs cannot be easily prompted to generate suspenseful stories is because suspense is an affective response to a cognitive state that only comes about under certain circumstances.
There must be a protagonist that is empathetic to the reader.
That protagonist must face a high possibility of an outcome undesired by the protagonist and the reader.
The quantity and quality (roughly in terms of expected probability) of ways of avoiding that undesired outcome are reduced.
This suggests an adversarial planning process wherein the generator establishes the conditions for suspense in three stages of generation.
\textbf{Background Setup} \S\ref{sec:background-setup} establishes the necessary story background, describing the protagonist, their goal, and a dire situation the protagonist will be put in if the goal is not successfully achieved.
\textbf{Outline Planning} \S\ref{sec:outline-planning} plans out the story outline by providing a series of possible actions for the protagonist to take to achieve their goal, and then generating potential reasons why actions might turn out to be ineffective in achieving the goal.
In essence, we craft a plan for the protagonist to escape the undesired outcome, and then adversarially establish the conditions that block the protagonist from that escape.
\textbf{Detail Elaboration} \S\ref{sec:detail-elaboration} composes event sequences that manifest each of the planned actions and failures, then elaborates on the details of the events.
Our method does not rely on any external story corpora.
Once the problem is decomposed in this way, the LLM is capable of generating all intermediate information required for generating a full suspenseful story.

\subsection{Background Setup}
\label{sec:background-setup}

We first sample a genre (e.g., spy thriller) from a list of suspenseful story genres\footnote{We provide the list in \autoref{sec:appendix-genre-list}.} obtained by prompting the LLM.
Next, we ask the LLM for the name and occupation of the protagonist.
Conditioned on that information, we prompt the LLM to design a goal for the protagonist to achieve.
Based on the goal, we then use the LLM to generate a dire situation that the protagonist would be put in, if the protagonist couldn't achieve the goal.
After that, we feed all the information to the LLM and ask it to generate an introductory paragraph of a story that can cover the given information.
The prompts we use in the background setup stage and an example of LLM's corresponding outputs are presented in \autoref{tab:background-setup}.

\begin{table}[!htb]
\centering
\small
\begin{tabular}{p{0.95\linewidth}}

\toprule

\textbf{Tell me about the name and the occupation of a protagonist of a spy thriller.} \\
\textit{Max Sterling, a highly skilled and enigmatic operative for a clandestine government agency tasked with eliminating high-profile threats to global security.} \\
(We denote the model output as \texttt{\$name\_occupation}.) \\

\midrule[0.03em]

\textbf{Tell me about a very concrete goal the protagonist wants to achieve.} \\
\textit{Max Sterling is determined to track down the elusive leader of an international terrorist organization known as ``The Phoenix'', who is planning a devastating attack on a major world summit. Max's goal is to gather enough evidence to dismantle the organization and bring their leader to justice before innocent lives are lost.} \\
(We denote the model output as \texttt{\$goal}.) \\

\midrule[0.03em]

\textbf{Tell me about a dire situation that the protagonist would be put in, if the protagonist couldn't achieve the goal.} \\
\textit{If Max Sterling fails to achieve their goal, ``The Phoenix'' will successfully carry out their planned attack, resulting in the loss of countless lives and substantial damage to global security. With the organization’s leader remaining at large, they would continue orchestrating acts of terrorism, further destabilizing the world and posing an ongoing threat that would be difficult to contain. Max understands the urgency and stakes involved, knowing that failure could lead to catastrophic consequences.} \\
(We denote the model output as \texttt{\$dire\_situation}.) \\

\midrule[0.03em]

\textbf{Write an introductory paragraph of a story that covers all the above information.} \\
\textit{In the shadows of a world teetering on the brink, an enigmatic operative named Max Sterling emerged from the depths of secrecy. Loyal to a clandestine government agency devoted to preserving global security, Max possessed an unmatched skillset and a steadfast determination. Their mission: to unearth and dismantle ``The Phoenix'', an international terrorist organization poised to unleash chaos upon a major world summit. As the clock ticked ominously, Max embarked on a relentless pursuit, driven by the weight of countless innocent lives hanging in the balance. Failure would mean catastrophic consequences, with ``The Phoenix'' thriving and global security faltering. But Max, undeterred by the daunting odds, knew that the time for redemption had come with each heartbeat. The fate of nations rested upon their resilient shoulders, and the chase had
only just begun.} \\

\bottomrule
\end{tabular}
\caption{The prompts we use in \autoref{sec:background-setup} Background Setup (highlighted in \textbf{bold}), and an example of LLM's corresponding outputs (highlighted in \textit{italic}). The genre of this story example is \textit{spy thriller}.}
\label{tab:background-setup}
\end{table}

\subsection{Outline Planning}
\label{sec:outline-planning}

Inspired by the psychological theory on suspense~\cite{gerrig1994readers}, we plan the story outline by iteratively creating story plots in which the protagonist takes actions trying to achieve the goal, followed by the failure of the actions.
Specifically, we realize it by iteratively prompting the LLM to (step \#1)~generate a series of possible actions the protagonist could take to achieve the goal.
For each action, we also prompt the LLM to (step \#2)~generate a potential reason that makes the action turn out to be ineffective in helping the protagonist achieve the goal---avoid the negative outcome---and get out of the dire situation.
To maintain the story coherence, during prompting we make sure that each action is generated conditioned on the failure of all previous actions, and that the likelihood of the series of actions is decreasing---readers assume a rational protagonist will try the best options first.
In this way, readers can start to feel suspense as they witness the gradual elimination of a series of feasible plans to avoid the negative outcome while the negative outcome draws closer and time runs out.

The prompts we use in the outline planning stage and an example of LLM's corresponding outputs are presented in \autoref{tab:outline-planning}.
For the illustration purpose, in this example we only perform three iterations of prompting and get three pairs of actions and reasons; in the third iteration we only keep \texttt{\$action3} and omit \texttt{\$reason3} as we want the protagonist to succeed at the end of the story.
In practice, there is no limitation on the number of prompting iterations, allowing our method to generate an unlimited length of stories with increased levels of storyline complexity.

\begin{table}
\centering
\small
\begin{tabular}{p{0.95\linewidth}}

\toprule

\textbf{Protagonist's Name and Occupation: {\texttt{\$name\_occupation}}} \\
\textbf{Protagonist's Goal: \texttt{\$goal}} \\
\textbf{Protagonist's Dire Situation (if the goal is not achieved): \texttt{\$dire\_situation}} \\
\textbf{Tell me about a concrete action the protagonist is most likely to take, in order to achieve the goal.} \\
\textit{Max Sterling, employing their extensive network of contacts and utilizing their advanced tech skills, infiltrates a top-secret meeting of high-ranking Phoenix members, planting surveillance devices to gather valuable intel on the organization’s plans and the whereabouts of its leader.} (We denote the model output as \texttt{\$action1}.) \\

\midrule[0.03em]

\textbf{The protagonist tries to perform the following action: \texttt{\$action1}.} \textbf{But the protagonist still fails to achieve the goal for a reason. Tell me about what this reason could be.} \\
\textit{Despite Max Sterling’s meticulous planning and execution, they are betrayed by a double agent within their own agency, who tips off the Phoenix leaders about the surveillance devices, resulting in the immediate evacuation of the meeting and thwarting Max’s attempt to gather crucial information and track down the organization’s leader.} (We denote the model output as \texttt{\$reason1}.) \\

\midrule[0.03em]

\textbf{The protagonist has taken all the following actions: \texttt{[\$action1]}.} \textbf{But it turns out that those actions are not effective in helping the protagonist to achieve the goal, so the goal is still not achieved yet. Tell me about a concrete new action the protagonist is most likely to take next, in order to achieve the goal.} \\
\textit{Max Sterling decides to track down and approach a former member of ``The Phoenix'', who defected years ago but still holds valuable information about the organization and its leader. Using their negotiating skills and persuasive tactics, Max aims to convince the defector to cooperate and provide crucial insider details that will help bring down the terrorist organization once and for all.} (We denote the model output as \texttt{\$action2}.) \\

\midrule[0.03em]

\textbf{The protagonist tries to perform the following action: \texttt{\$action2}.} \textbf{But the protagonist still fails to achieve the goal for a reason. Tell me about what this reason could be.} \\
\textit{Despite Max’s best efforts, the defector is not willing to collaborate, fearing for their own safety and the safety of their loved ones. The defector is haunted by past experiences with ''The Phoenix'' and their leader, making them reluctant to trust anyone, even someone as skilled and determined as Max.} (We denote the model output as \texttt{\$reason2}.) \\

\midrule[0.03em]

\textbf{The protagonist has taken all the following actions: \texttt{[\$action1, \$action2]}.} \textbf{But it turns out ... achieve the goal.} (similar to the prompt used for \texttt{\$action2}) \\
\textit{Realizing that their previous actions have not yielded the desired results, Max Sterling decides to gain access to The Phoenix’s encrypted communication channels, using their expertise in cryptography and hacking to unravel the organization’s hidden communications and identify potential weak links within the group who could provide vital information about the leader’s location.} (We denote the model output as \texttt{\$action3}.) \\

\bottomrule
\end{tabular}
\caption{The prompts we use in \autoref{sec:outline-planning} Outline Planning (highlighted in \textbf{bold}), and an example of LLM's corresponding outputs (highlighted in \textit{italic}). All the variables shaped as \texttt{\$variable} are obtained in \autoref{sec:background-setup}, as reflected in \autoref{tab:background-setup}.
}
\label{tab:outline-planning}
\end{table}

\subsection{Detail Elaboration}
\label{sec:detail-elaboration}

With the action-reason pairs obtained in \autoref{sec:outline-planning}, we convert each pair into a short summary for a story chapter in which the protagonist tries to take the given action but that action does not work due to the given reason.
Conditioned on the short summary, we prompt the LLM to generate a sequence of events that can serve as an outline for each story chapter.
During prompting we need to make sure the protagonist is not aware that their plan will fail (readers assume the protagonist is rational and will not take actions that they believe are unlikely to be effective in achieving their goal).
With a sequence of events, the full story is then generated by asking the LLM for an elaboration on the event details.

Aiming to further improve the level of suspense readers could possibly perceive, we incorporate two optional designs into our method for suspenseful story generation.
(1)~we ask the LLM to set up some small clues for the upcoming failure of the protagonist’s action so as to better arouse readers’ anticipation of the protagonist’s failure.
(2)~We draw inspiration from the narrative theory of {\em disparity of knowledge}~\cite{branigan2013narrative} and design two information revelation modes.
In the first, we reveal the reason the protagonist's actions will fail {\em beforehand}, setting up an anticipation of failure in the reader that the protagonist doesn't know about and is thus unlikely to be able to avert.
For example, if our protagonist is a spy, they might be unaware that the antagonists have set a trap.
In the second, we reveal the information {\em after the fact}, explaining to the reader how the protagonist's actions were doomed to fail from the beginning.

A complete run of our method starts with \S\ref{sec:background-setup} Background Setup and may contain two or more prompting iterations. Each of the prompting iterations comprises \S\ref{sec:outline-planning} Outline Planning and \S\ref{sec:detail-elaboration} Detail Elaboration, which are repeated in our method to create stories in which the protagonist repeatedly fails to prevent the dire situation.
In the final iteration, we do not generate the failure response (i.e., step \#2 of \autoref{sec:outline-planning}), and we allow the protagonist to succeed in preventing the dire situation and achieving the goal.
The prompts we use in the detail elaboration stage and an example of LLM's corresponding outputs are presented in \autoref{sec:appendix-prompt-for-detail-elaboration-event-sequence}. The corresponding full story is presented in \autoref{sec:appendix-full-story}.

\section{Evaluation}

Human participant evaluation has been generally held as the gold-standard evaluation technique for creative language generation tasks such as story generation~\cite{celikyilmaz2020evaluation,caglayan-etal-2020-curious,van2021human}.
To extensively evaluate our method, we design three human participant studies.
In the first, participants are asked to judge the goodness of the stories we generate and compare stories generated by our technique against those generated by the baseline method.
The comparisons are based on multiple evaluation criteria.
In the second, we perform ablations on our system and compare our full system against the ablations.
Ablation studies help us understand whether different parts of our system are significant to the solution.
In the third, participants are requested to examine the story outline to help determine if the internal workflow is running properly as we expected.

All human participant studies are performed using the Prolific crowdsourcing platform\footnote{\url{https://www.prolific.co/}}.
These studies have been approved by our institution’s Institutional Review Board (IRB).
We qualify participants by first asking them a screening question at the beginning of the survey, and then verifying answers manually to disregard answers provided by those who fail the screening.
We require participants to be physically located in the U.S. and to speak English as a first language.
For each human participant study, we source a distinct set of participants (no overlap) to avoid potential bias in annotations that could occur from participating in related studies in the past.
For each study, we measure the average inter-annotator agreement using Fleiss's kappa~\cite{fleiss1971measuring}.

We choose ChatGPT~\cite{openai2022chatgpt} (\texttt{gpt-3.5-turbo-0613}) as the LLM we use in the implementation.
As mentioned in \autoref{sec:method}, our method does not rely on any external story corpora.
Suspenseful stories can be generated with our method in a fully zero-shot way from ChatGPT.
We start every ChatGPT dialogue session with ``You are a creative storyteller.''.
When generating anything other than stories (e.g., interim information such as the protagonist's goal and actions), to discourage ChatGPT from generating excessively long responses that are usually not quite useful when serving as inputs to subsequent generation stages, we append ``Use no more than [$n$] sentences in your answer.'' ($n$ can be three, four, or five) to the prompt when necessary.

\subsection{Comparison Against Baseline Method}
\label{sec:main-result}

In this study, we seek to understand whether our method has achieved improvement in terms of story suspense as well as other typical story evaluation criteria.
We choose the following method as the baseline to compare with: directly asking ChatGPT for a suspenseful story with straightforward prompts.
To make it a fair comparison, we control that the stories are generated conditioned on the same background information (the protagonist's name, occupation, and goal) used in our method's generation process. The prompt we use for generating the baseline stories is: ``Story Background: \texttt{\$name\_occupation}. \texttt{\$goal}. Write a full suspenseful story based on the story background.'' (variables \texttt{\$name\_occupation} and \texttt{\$goal} are obtained from \autoref{sec:background-setup}).

We recruited 90 participants to evaluate 30 pairs of stories, with each individual evaluating 10 pairs of stories, and each pair of stories was evaluated by 30 participants.
Each pair of stories comprises one story generated by our method and one story generated by the baseline method.
Both stories are randomly selected from a corresponding large collection of stories generated with either our method or the baseline method.

We ask human participants to make a pair-wise comparison and assess the goodness of pairs of suspenseful stories from the following dimensions:
\begin{itemize}[noitemsep]
    \setlength\itemsep{0em}
    \item \textbf{Suspense:} ``Which story makes you feel a higher level of suspense?'';
    \item \textbf{Novelty:} ``Which story do you find to be more novel?'';
    \item \textbf{Enjoyment:} ``Which story is more enjoyable to read?'';
    \item \textbf{Logical Sense:} ``Which story logically makes more sense?'';
    \item \textbf{Naturalness:} ``Which story is more likely to have been written by a human?''.
\end{itemize}

\begin{table*}
\centering
\resizebox{\linewidth}{!}{
\begin{tabular}{c|lll|lll|lll|lll|lll}
    \toprule
    \multirow{2}{*}{\textbf{Methods}}
    & \multicolumn{3}{c|}{\textbf{Suspense}} & \multicolumn{3}{c|}{\textbf{Novelty}} & \multicolumn{3}{c|}{\textbf{Enjoyment}} & \multicolumn{3}{c|}{\textbf{Logical Sense}} & \multicolumn{3}{c}{\textbf{Naturalness}}
    \\
    &Win\%&Lose\%&Tie\%&Win\%&Lose\%&Tie\%&Win\%&Lose\%&Tie\%&Win\%&Lose\%&Tie\%&Win\%&Lose\%&Tie\% \\
    
    \midrule
    
    Ours v.s. Baseline
    & \textbf{84.9}** & 11.8 & 3.3
    & \textbf{76.6}** & 13.6 & 9.8
    & \textbf{67.9}** & 19.4 & 12.7
    & \textbf{49.2}* & 27.8 & 23.0
    & 25.3 & 18.6 & \textbf{56.1}**
    \\

    \midrule
    \midrule
    
    Ours v.s. Ablation \#1
    & \textbf{75.0}** & 19.2 & 5.8
    & \textbf{63.1}** & 16.5 & 20.4
    & \textbf{59.9}** & 17.2 & 22.9
    & \textbf{37.8} & 27.1 & 35.1
    & 31.0 & 29.9 & \textbf{39.1}
    \\
    Ablation \#1 v.s. Baseline
    & \textbf{79.8}** & 12.9 & 7.3
    & \textbf{68.9}** & 11.3 & 19.8
    & \textbf{71.7}** & 12.1 & 16.2
    & \textbf{45.3} & 30.9 & 23.8
    & 33.2 & 28.6 & \textbf{38.2}
    \\

    \midrule
    
    Ours v.s. Ablation \#2
    & \textbf{80.4}** & 9.9 & 9.7
    & \textbf{66.0}** & 10.3 & 23.7
    & \textbf{68.6}** & 16.6 & 14.8
    & \textbf{39.4} & 32.4 & 28.2
    & 22.8 & 24.0 & \textbf{53.2}**
    \\
    Ablation \#2 v.s. Baseline
    & \textbf{72.3}** & 12.7 & 15.0
    & \textbf{74.3}** & 9.9 & 15.8
    & \textbf{63.1}** & 15.3 & 21.6
    & 29.6 & 22.3 & \textbf{48.1}*
    & 30.9 & 25.4 & \textbf{43.7}
    \\
    
    \bottomrule
\end{tabular}
}
\caption{Human evaluation results for both the comparison against the ChatGPT baseline and the ablation studies, showing the percentage of times stories generated from each method are preferred for five evaluation metrics. ``Win'' means the method on the left is preferred. ** indicates results are significant at $p<0.01$ (* at $p<0.05$) confidence level using a Wilcoxon sign test.}
\label{tab:ours-vs-baseline}
\end{table*}

\begin{table*}
\centering
\resizebox{\linewidth}{!}{
\begin{tabular}{c|lll|lll|lll|lll|lll}
    \toprule
    \multirow{2}{*}{\textbf{Methods}}
    & \multicolumn{3}{c|}{\textbf{Suspense}} & \multicolumn{3}{c|}{\textbf{Novelty}} & \multicolumn{3}{c|}{\textbf{Enjoyment}} & \multicolumn{3}{c|}{\textbf{Logical Sense}} & \multicolumn{3}{c}{\textbf{Naturalness}}
    \\
    &Win\%&Lose\%&Tie\%&Win\%&Lose\%&Tie\%&Win\%&Lose\%&Tie\%&Win\%&Lose\%&Tie\%&Win\%&Lose\%&Tie\% \\
    
    \midrule
    
    Ours v.s. Baseline
    & \textbf{73.9}** & 15.7 & 10.4
    & \textbf{66.2}** & 9.6 & 24.2
    & \textbf{59.8}** & 18.3 & 21.9
    & \textbf{42.6} & 23.6 & 33.8
    & 25.1 & 19.9 & \textbf{55.0}**
    \\
    
    \bottomrule
\end{tabular}
}
\caption{Human evaluation results for the comparison against the Llama 2 baseline, showing the percentage of times stories generated from each method are preferred for five evaluation metrics. ``Win'' means the method on the left is preferred. ** indicates results are significant at $p<0.01$ confidence level using a Wilcoxon sign test.}
\label{tab:llama2}
\end{table*}

\begin{table*}
\centering
\resizebox{\linewidth}{!}{
\begin{tabular}{c|lll|lll|lll|lll|lll}
    \toprule
    \multirow{2}{*}{\textbf{Methods}}
    & \multicolumn{3}{c|}{\textbf{Suspense}} & \multicolumn{3}{c|}{\textbf{Novelty}} & \multicolumn{3}{c|}{\textbf{Enjoyment}} & \multicolumn{3}{c|}{\textbf{Logical Sense}} & \multicolumn{3}{c}{\textbf{Naturalness}}
    \\
    &Win\%&Lose\%&Tie\%&Win\%&Lose\%&Tie\%&Win\%&Lose\%&Tie\%&Win\%&Lose\%&Tie\%&Win\%&Lose\%&Tie\% \\
    
    \midrule
    
    Ours v.s. Baseline
    & \textbf{81.1}** & 6.8 & 12.1
    & \textbf{84.7}** & 11.3 & 4.0
    & \textbf{70.8}** & 11.9 & 17.3
    & 33.1 & 25.7 & \textbf{41.2}
    & 28.4 & 12.0 & \textbf{59.6}**
    \\
    
    \bottomrule
\end{tabular}
}
\caption{Human evaluation results for the comparison against the Re3 baseline, showing the percentage of times stories generated from each method are preferred for five evaluation metrics. ``Win'' means the method on the left is preferred. ** indicates results are significant at $p<0.01$ confidence level using a Wilcoxon sign test.}
\label{tab:re3}
\end{table*}

Results are presented in \autoref{tab:ours-vs-baseline} where we show the percentage of times stories generated from each method are preferred for each evaluation metric.
Participants' preferences with a fair average inter-annotator agreement show that compared to the baseline, our method significantly improves the human perception of narrative suspense, and effectively produces more novel and enjoyable stories.
Our method also achieves improvement in the logical sense of stories.
We attribute this to our method's outline planning component which can make the narratives more rational and reasonable.
As our method and the baseline method both use ChatGPT, the naturalness of our stories stays at the same level as the baseline's.

To further validate the effectiveness of our method across different LLMs, we perform an additional experiment on Llama 2~\cite{touvron2023llama} (\texttt{Llama-2-13b-chat}). We follow the same experimental setup as introduced before, except changing the LLM from ChatGPT to Llama 2. Results presented in \autoref{tab:llama2} show that the Llama 2 augmented with our method can generate more suspenseful, more novel, and more enjoyable stories compared to a vanilla Llama 2. These results echo the main results reported in \autoref{tab:ours-vs-baseline} and therefore prove that our method is stable and transferrable across different LLMs.

Since the stories generated by our method are long-form (see \autoref{sec:appendix-full-story} for an example), we include Re3~\cite{yang-etal-2022-re3}, a commonly used long-form story generation technique, as an additional baseline to compare with. We condition both Re3 and our method on the exact same background information to generate full stories. Results are presented in \autoref{tab:re3}, showing that our method can generate more suspenseful, more novel, and more enjoyable stories compared to Re3.

\subsection{Ablation Studies}

To fully investigate the effectiveness of our method and confirm the necessity of two components: outline planning (\S\ref{sec:outline-planning}) and detail elaboration (\S\ref{sec:detail-elaboration}), we perform two corresponding ablation studies.
First, to study the necessity of detail elaboration, we ablate our method (denoted as Ablation \#1) by replacing the detail elaboration component with a straightforward prompt as presented in \autoref{tab:prompt-ablation}.
Second, to study the necessity of outline planning, based on Ablation \#1 we ablate our method (denoted as Ablation \#2) by further replacing the outline planning component with a similarly straightforward prompt (also listed in \autoref{tab:prompt-ablation}).
We recruited 90 participants and followed the evaluation setup of \autoref{sec:main-result}.
Participants' preferences presented in \autoref{tab:ours-vs-baseline} with a fair average inter-annotator agreement show that both outline planning (\S\ref{sec:outline-planning}) and detail elaboration (\S\ref{sec:detail-elaboration}) are necessary for generating novel and enjoyable suspenseful stories.

\begin{table}[t]
\centering
\small
\begin{tabular}{p{0.95\linewidth}}

\toprule

Ablation \#1's Prompt \\

\midrule[0.03em]

\textbf{Story Background: \texttt{\$name\_occupation}. \texttt{\$goal}.} \\
\textbf{Action \#1: \texttt{\$action1}} \\
\textbf{Reason \#1: \texttt{\$reason1}} \\
\textbf{Action \#2: \texttt{\$action2}} \\
\textbf{Reason \#2: \texttt{\$reason2}} \\
\textbf{Action \#3: \texttt{\$action3}} \\
\textbf{Story Summary: Aiming to achieve the goal: \texttt{\$goal}, the protagonist first tries to take Action \#1, but fails due to Reason \#1; then tries to take Action \#2, but fails due to Reason \#2; finally tries to take Action \#3 and this time the protagonist successfully achieves the goal.} \\
\textbf{Write a full suspenseful story based on the story summary.} \\

\midrule

Ablation \#2's Prompt \\

\midrule[0.03em]

\textbf{Story Background: \texttt{\$name\_occupation}. \texttt{\$goal}.} \\
\textbf{Story Summary: Aiming to achieve the goal: \texttt{\$goal}, the protagonist first tries to take a first action, but fails due to a reason; then tries to take another new action, but fails due to another reason; finally tries to take a new action and this time the protagonist successfully achieves the goal.} \\
\textbf{Write a full suspenseful story based on the story summary.} \\

\bottomrule
\end{tabular}
\caption{The prompts we use in ablation studies. All the variables shaped as \texttt{\$variable} are obtained in \autoref{sec:background-setup} and \autoref{sec:outline-planning}, as reflected in \autoref{tab:background-setup} and \autoref{tab:outline-planning}.
}
\label{tab:prompt-ablation}
\end{table}

\subsection{Examination of Story Outline}

After confirming the effectiveness of our method and the necessity of individual method components, we need to further examine the story outline generated in \autoref{sec:outline-planning} to determine if our method's internal workflow is running properly as we expected.
We asked (1)~whether the actions of the protagonist are relevant to the goal, (2)~whether a plausible reason is given for the protagonist's failure, and (3)~whether each of the protagonist's actions, after each failure, is perceived as decreasing in likelihood.

For each aspect, we recruited 30 participants to examine the goal, actions, and reasons for failure of 30 randomly selected stories generated by our method.
On the question of {\bf protagonist action related to goal}, participants reported that protagonist actions were relevant to the goal 96.5\% of the time.
Regarding {\bf plausible reasons for failure}, 89.8\% of the time.
We find that in the vast majority of cases, the action is relevant to the goal and the reason is plausible for the protagonist's failure.
On the question of {\bf decreasing likelihood}, 55.1\% of participants indicated that they thought the likelihood of actions was going down, 13.9\% indicated that they thought the likelihood was going up, and 31.0\% thought the likelihood stayed the same.
More than half of participants perceived declining likelihood trends, and a clear plurality felt that likelihoods were decreasing or staying the same.
These results all indicate that the internal workflow of our method is running properly as we expected.

\section{Factors Affecting Suspense Perception}

Given the success of our method in creating suspenseful stories, we recognized a unique opportunity to contribute to a more fine-grained understanding of suspense.
Prior work from psychology used hand-crafted stories from literature and thus did not have the means to run controlled studies with a large number of stories with fine-grained interventions.
We conducted a study to investigate the timing and ordering of knowledge revealed to the reader, as well as the relationship between empathy for the protagonist and suspense.
We specifically investigate how the following affect suspense perception:
(1)~clues that set up the failure;
(2)~information revelation timing (before vs after);
(3)~reader's empathy.

\textbf{Clue setup}: we recruited 90 participants to ask for their preferences regarding the suspense level of 30 randomly selected pairs of stories.
For each pair of stories, which is generated with and without clue setup (detailed in \autoref{sec:detail-elaboration}) respectively, we ask participants which one of the two stories makes them feel a higher level of suspense.
57.9\% of participants reported higher suspense when reading stories with clues, whereas 10.9\% of participants reported higher suspense when reading stories without clues (31.2\% were a tie).
These results are significant at $p < 0.05$ confidence level using a Wilcoxon sign test.
Thus we conclude that clues of impending failure are a factor that increases reader perceptions of suspense.

\textbf{Information revelation timing}: we recruited 90 participants to ask for their preferences regarding the suspense level of 30 randomly selected pairs of stories.
For each pair of stories, which is generated by revealing the information \textit{beforehand} and \textit{after the fact} (detailed in \autoref{sec:detail-elaboration}) respectively, we ask participants which one of the two stories makes them feel a higher level of suspense.
42.1\% of participants reported higher suspense when reading a story that explained the impending failure beforehand, and 36.3\% of participants reported higher suspense when the impending failure was explained after the fact (21.6\% tied).
This indicates that it is slightly more effective in facilitating reader suspense perceptions by explaining how the protagonist will fail beforehand than after the fact.
These results are not statistically significant.

\begin{figure}[t]
\centering
\includegraphics[width=\linewidth]{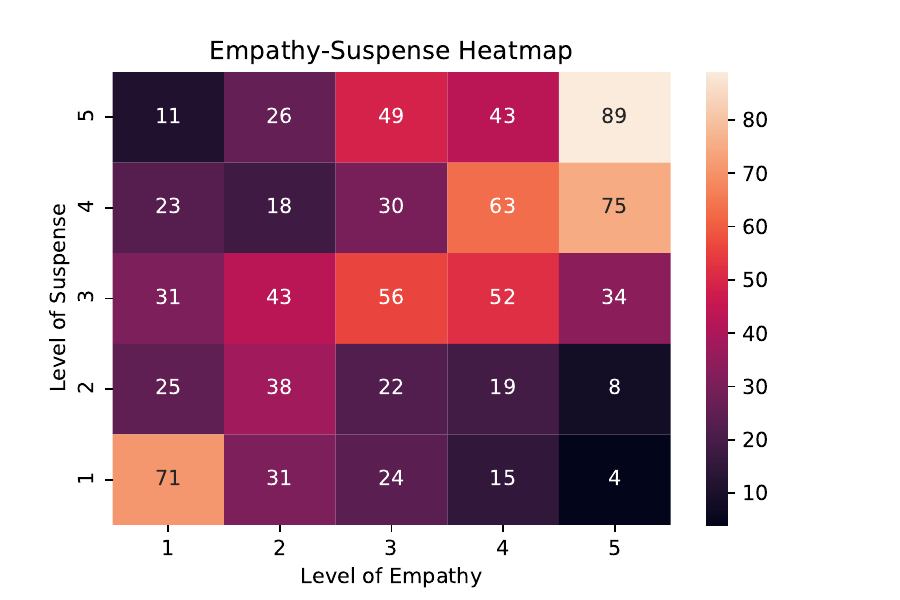}
\caption{Human evaluation results for the investigation on reader's empathy as visualized in a heatmap. The numbers in the heatmap indicate the counts of cases in which human participants feel the corresponding levels of empathy and suspense after reading a suspenseful story.}
\label{fig:heatmap}
\end{figure}

\textbf{Reader empathy}: we recruited 30 participants to ask for both the level of empathy for the protagonist and the level of suspense they feel (level=1/2/3/4/5) after reading 30 randomly selected suspenseful stories generated by our method.
Results are presented in \autoref{fig:heatmap} as a heatmap which shows that a high level of perceived suspense is usually associated with a high level of reader's empathy.
However, there does exist a noticeable portion of participants who exhibit a very low level of perceived suspense even when they are reading suspenseful stories, associated with a very low level of empathy.
One possible explanation for this phenomenon is methodological.
Not a lot of character setup is given in the experiments and empathy may not be triggered strongly in some participants, who then also do not perceive suspense.

\section{Conclusions}

In this paper, we propose a theory-grounded \textit{iterative-prompting-based planning} approach to generating suspenseful stories.
To the best of our knowledge, this is the first attempt at suspenseful story generation with LLMs.
Through extensive human evaluations, we prove that our method is more capable of producing suspenseful stories than the baseline method.
Our in-depth analysis of factors that affect suspense perception is also informative for story generation researchers and practitioners.

\section{Limitations}

Our system is developed for suspenseful story generation in English and our method has not been tested in other languages.
The method is grounded in theory based on analysis of stories from a Western storytelling tradition.
We believe that our method intuition is directly transferrable to other language scenarios, but to develop a fully functional non-English suspenseful story generation system, much additional effort must be required, especially for the necessary prompt engineering and human evaluations.

\section{Ethical Considerations}

This work uses human participants for evaluation. All procedures in this work have been approved by our institution’s Institutional Review Board (IRB). We provide the following statement on the first page of all our surveys, to make the human participants aware of the potential risk: ``This survey may contain sensitive, distressing, or potentially triggering content. Please proceed only if you feel comfortable and are prepared to engage with thriller stories.''.

We ensure the human participants are fairly compensated by paying them at (and sometimes above) the hourly rate recommended by Prolific. We reasonably estimate the survey completion time to be two or three times (depending on the survey's difficulty level we perceive) the time we have spent taking the survey.

\bibliography{anthology,custom}

\appendix

\section{List of Suspenseful Story Genres}
\label{sec:appendix-genre-list}
\begin{itemize}
    \item Psychological Thriller
    \item Crime Thriller
    \item Mystery Thriller
    \item Action Thriller
    \item Supernatural Thriller
    \item Survival Thriller
    \item Conspiracy Thriller
    \item Horror Thriller
    \item Political Thriller
    \item Survival Horror
    \item Techno Thriller
    \item Espionage Thriller
    \item Neo-Noir Thriller
    \item Erotic Thriller
    \item Disaster Thriller
    \item Serial Killer Thriller
    \item Home Invasion Thriller
    \item Legal Thriller
    \item Medical Thriller
    \item Historical Thriller
    \item Science Fiction Thriller
    \item War Thriller
    \item Spy Thriller
    \item Revenge Thriller
    \item Adventure Thriller
    \item Gothic Thriller
    \item Slasher Thriller
    \item Found Footage Thriller
    \item Post-Apocalyptic Thriller
    \item Wilderness Thriller
    \item Cyber Thriller
    \item Survivalist Thriller
    \item Mind-Bending Thriller
    \item Mind Game Thriller
    \item Vigilante Thriller
    \item Martial Arts Thriller
    \item Chase Thriller
    \item Identity Thriller
    \item Betrayal Thriller
    \item Dark Comedy Thriller
    \item Heist Thriller
    \item Political Conspiracy Thriller
    \item Stalker Thriller
    \item Paranormal Thriller
    \item Spy Action Thriller
    \item Neo-Western Thriller
    \item Psychological Horror Thriller
    \item Gothic Horror Thriller
    \item Found Footage Horror Thriller
    \item Survival Drama Thriller
\end{itemize}

\newpage
\onecolumn
\section{Prompts for Detail Elaboration and An Example of Event Sequence}
\label{sec:appendix-prompt-for-detail-elaboration-event-sequence}
\begin{small}
\begin{longtable}{p{0.9\textwidth}}

\toprule

\textbf{Elaborate the following story summary into a story chapter by writing a sequence of events for me.} \\
\textbf{Make sure the protagonist is not aware of the reason until the action is taken.} \\
\textbf{Make sure to reveal the reason for the action's ineffectiveness in late stages of the story.} \\
\textbf{Make sure to set up some small clues that can suggest the protagonist's action's ineffectiveness.} \\
\textbf{Story Summary:} \textbf{The protagonist tries to take the following action: \texttt{\$action1}.} \textbf{But it does not work due to the following reason: \texttt{\$reason1}.} \\
\textbf{Story Chapter (i.e. a sequence of events):} \\
\textit{\textbullet~Max Sterling, a skilled operative with an extensive network of contacts and advanced tech skills, meticulously planned to infiltrate a top-secret meeting of high-ranking Phoenix members. The goal was to plant surveillance devices that would gather valuable intel on the organization's plans and the whereabouts of its elusive leader.} \\
\textit{\textbullet~Max spent weeks gathering intelligence, studying the layout of the meeting location, and recruiting fellow agents to assist with the operation. The night of the mission arrived, and Max, disguised as a maintenance worker, slipped into the heavily guarded compound unnoticed.} \\
\textit{\textbullet~With precise timing, Max managed to slip into the room where the meeting was taking place. Using their expert knowledge of technology, Max discreetly planted the surveillance devices in strategic locations, making sure they were hidden from sight. Everything seemed to be going according to plan.} \\
\textit{\textbullet~However, despite Max's caution and thorough preparations, a feeling of unease began to creep in. There was something off about the atmosphere in the room. Unbeknownst to Max, a double agent within their own agency had betrayed them. This agent, corrupted by the Phoenix organization, had tipped off the leaders about the surveillance devices.} \\
\textit{\textbullet~As Max silently made their way out of the room, a sudden commotion erupted. The Phoenix members became aware of the planted devices and realized they were under surveillance. Panic filled the air as they quickly evacuated the premises.} \\
\textit{\textbullet~Outside the meeting room, Max's heart sank. Their painstaking efforts had been in vain. The operation had been compromised, and the chance to gather crucial information about the organization now seemed lost.} \\
\textit{\textbullet~Confusion and frustration clouded Max's mind as they tried to comprehend what had just happened. How had they been betrayed? Who was the double agent within their own agency? Questions swirled, but answers seemed distant.} \\

\midrule[0.03em]

\textbf{Elaborate the following story summary into a story chapter by writing a sequence of events for me.} \\
\textbf{Make sure the protagonist is not aware of the reason until the action is taken.} \\
\textbf{Make sure to reveal the reason for the action's ineffectiveness in late stages of the story.} \\
\textbf{Make sure to set up some small clues that can suggest the protagonist's action's ineffectiveness.} \\
\textbf{Story Summary:} \textbf{The protagonist tries to take the following action: \texttt{\$action2}.} \textbf{But it does not work due to the following reason: \texttt{\$reason2}.} \\
\textbf{Story Chapter (i.e. a sequence of events):} \\
\textit{\textbullet~Max Sterling, a skilled agent with a relentless determination, had made it his mission to bring down the notorious terrorist organization, \"The Phoenix.\" Aware of a former member who had defected years ago, Max saw an opportunity to obtain valuable insider information that could cripple the organization once and for all.} \\
\textit{\textbullet~Determined to convince the defector to cooperate, Max embarked on a dangerous journey to track them down. Through countless hours of research, he pieced together clues and managed to locate the defector's current whereabouts.} \\
\textit{\textbullet~With a plan in mind, Max set out to approach the defector. He knew that his negotiating skills and persuasive tactics would be crucial in convincing them to share their knowledge. Max believed that with the right words, he could offer them a renewed sense of purpose and safety.} \\
\textit{\textbullet~As Max arrived at the defector's secluded hideout, a hint of nervous anticipation filled the air. He prepared himself mentally, going over the details of their previous affiliation with \"The Phoenix\" and the potential benefits of collaborating. Max believed that once the defector understood the impact their cooperation could have on dismantling the organization, they would surely be swayed.} \\
\textit{\textbullet~However, as Max began his carefully crafted speech, he quickly realized that it would be an uphill battle. The defector, haunted by their past experiences, was a broken soul. Their eyes reflected a deep-rooted fear, a visceral apprehension, that Max had not anticipated. They hesitated, their gaze filled with a mixture of fear and uncertainty.} \\
\textit{\textbullet~Despite Max's best efforts, the defector shook their head, unwilling to collaborate. They spoke of the constant threat \"The Phoenix\" posed to their own safety and that of their loved ones. Their voice quivered as they recalled the horrors they had witnessed under the organization's reign. To the defector, trusting anyone, even someone as skilled and determined as Max, seemed like an impossible feat.} \\
\textit{\textbullet~Max felt a twinge of disappointment, realizing the immense challenge he faced. He knew he couldn't force the defector to cooperate, and he respected their decision. It was clear that the wounds inflicted by \"The Phoenix\" ran deep, leaving scars that would not easily fade away.} \\
\textit{\textbullet~As he left the hideout, Max couldn't help but question his approach. He had hoped that his determination and persuasive techniques would be enough. Yet, he realized that sometimes, the damage caused by an organization's cruelty is too debilitating, leaving individuals unable to trust and risking their own lives and the lives of their loved ones in the process.} \\

\midrule[0.03em]

\textbf{Elaborate the following story summary into a story chapter by writing a sequence of events for me.} \\
\textbf{Make sure the protagonist is not aware of the reason until the action is taken.} \\
\textbf{Make sure to reveal the reason for the action's ineffectiveness in late stages of the story.} \\
\textbf{Make sure to set up some small clues that can suggest the protagonist's action's ineffectiveness.} \\
\textbf{Story Summary:} \textbf{The protagonist takes the following action: \texttt{\$action3}.} \textbf{Therefore, the protagonist successfully achieves the goal: \texttt{\$goal}.}
\textbf{The story ends with the protagonist's success.} \\
\textbf{Story Chapter (i.e. a sequence of events):} \\
\textit{\textbullet~Max Sterling had dedicated his life to hunting down the shadowy figure who led the international terrorist organization known as "The Phoenix." Their crimes had left a trail of devastation and despair around the world, and Max was determined to bring their reign of terror to an end.} \\
\textit{\textbullet~Frustration had gnawed at Max as his previous efforts had failed to yield any significant results. The Phoenix's encrypted communication channels remained impenetrable, shrouding their nefarious plans in secrecy. Max, however, possessed a unique skill set—a mastery of cryptography and hacking—that set him apart from the rest.} \\
\textit{\textbullet~Realizing that the clock was ticking and the world summit was drawing near, Max knew he had to take drastic action. In a dimly lit room filled with the soft hum of computers, he embarked on a relentless mission to breach The Phoenix's defenses.} \\
\textit{\textbullet~His fingers danced across the keyboard, his mind working in perfect harmony with the lines of code on the screen. It was a battle of wits as Max faced off against the organization's tech-savvy members. Layers of encryption were peeled back, firewalls breached, and digital traps skillfully avoided. Max's determination was unwavering.} \\
\textit{\textbullet~Days turned into nights as he delved deeper into the abyss of The Phoenix's digital world. The stakes were higher than ever, with the world summit fast approaching and the specter of a devastating attack looming. Failure was not an option.} \\
\textit{\textbullet~As Max progressed, he began to unravel The Phoenix's hidden communications. Messages, once obscured by complex algorithms, became clear. He identified potential weak links within the organization, individuals who could hold the key to unraveling the leader's identity and location.} \\
\textit{\textbullet~Each breakthrough brought him closer to his goal, but it also escalated the danger. The Phoenix, aware of Max's relentless pursuit, fought back with ferocity. They launched counterattacks, attempting to trace Max's location and expose him to their deadly network.} \\
\textit{\textbullet~But Max was relentless. He gathered a treasure trove of information about the organization's structure, their plans, and the identities of their key operatives. The evidence he collected was damning, and he knew it would be the key to dismantling The Phoenix once and for all.} \\
\textit{\textbullet~Finally, the moment of truth arrived. Max uncovered the location of The Phoenix's leader, a breakthrough that sent shivers down his spine. He had gathered enough evidence to not only bring down the organization but also to ensure the leader faced justice for their crimes.} \\
\textit{\textbullet~With the weight of the world on his shoulders, Max Sterling had succeeded in his mission. He had tracked down the elusive leader of The Phoenix and gathered the evidence needed to dismantle the organization. The world summit was safe, and innocent lives were spared from the impending disaster.} \\
\textit{\textbullet~As Max watched the authorities apprehend The Phoenix's leader, he knew that justice would prevail. The darkness that had shrouded The Phoenix was finally lifted, and a sense of closure washed over Max. His relentless pursuit of the truth had paid off, and he had achieved his mission to protect the world from the horrors of terrorism.} \\

\bottomrule
\caption{The prompts we use in \autoref{sec:detail-elaboration} Detail Elaboration (highlighted in \textbf{bold}), and an example of LLM's corresponding outputs (highlighted in \textit{italic}). All the variables shaped as \texttt{\$variable} are obtained in \autoref{sec:background-setup} and \autoref{sec:outline-planning}, as reflected in \autoref{tab:background-setup} and \autoref{tab:outline-planning}.
}
\label{tab:detail-elaboration}
\end{longtable}
\end{small}

\newpage
\onecolumn
\section{An Example of Full Story}
\label{sec:appendix-full-story}
\textbf{Introductory paragraph:} In the shadows of a world teetering on the brink, an enigmatic operative named Max Sterling emerged from the depths of secrecy. Loyal to a clandestine government agency devoted to preserving global security, Max possessed an unmatched skillset and a steadfast determination. Their mission: to unearth and dismantle \"The Phoenix,\" an international terrorist organization poised to unleash chaos upon a major world summit. As the clock ticked ominously, Max embarked on a relentless pursuit, driven by the weight of countless innocent lives hanging in the balance. Failure would mean catastrophic consequences, with \"The Phoenix\" thriving and global security faltering. But Max, undeterred by the daunting odds, knew that the time for redemption had come with each heartbeat. The fate of nations rested upon their resilient shoulders, and the chase had only just begun.

\textbf{Chapter \#1:} Max Sterling, a skilled operative with an extensive network of contacts and advanced tech skills, meticulously planned to infiltrate a top-secret meeting of high-ranking Phoenix members. The goal was to plant surveillance devices that would gather valuable intel on the organization's plans and the whereabouts of its elusive leader. Max spent weeks gathering intelligence, studying the layout of the meeting location, and recruiting fellow agents to assist with the operation. The night of the mission arrived, and Max, disguised as a maintenance worker, slipped into the heavily guarded compound unnoticed. The atmosphere was charged with tension as Max's determination pushed them onward. With precise timing, Max managed to slip into the room where the meeting was taking place. Using their expert knowledge of technology, Max discreetly planted the surveillance devices in strategic locations, making sure they were hidden from sight. Everything seemed to be going according to plan. However, despite Max's caution and thorough preparations, a feeling of unease began to creep in. There was something off about the atmosphere in the room. Unbeknownst to Max, a double agent within their own agency had betrayed them. This agent, corrupted by the Phoenix organization, had tipped off the leaders about the surveillance devices. As Max silently made their way out of the room, a sudden commotion erupted. The Phoenix members became aware of the planted devices and realized they were under surveillance. Panic filled the air as they quickly evacuated the premises. Max's heart sank. Outside the meeting room, Max's mind raced trying to make sense of the situation. Their painstaking efforts had been in vain. The operation had been compromised, and the chance to gather crucial information about the organization now seemed lost. The traitor within the agency had jeopardized not only the mission but also the safety of countless lives. Confusion and frustration clouded Max's mind as they tried to comprehend what had just happened. How had they been betrayed? Who was the double agent within their own agency? Questions swirled, but answers seemed distant. Max needed to regroup and find a new strategy to bring down The Phoenix. Over the next few days, Max delved deep into investigation mode. They meticulously analyzed every detail of the failed operation, reviewed intelligence reports, and interviewed potential suspects within the agency. With each piece of evidence, Max narrowed down the potential double agents, determined to uncover the truth. As Max got closer to unveiling the identity of the double agent, danger loomed. The Phoenix organization had caught wind of Max's relentless pursuit and sent their assassins after them. Max was forced to constantly stay on the move, using their extensive network of contacts to gather information and evade capture. Finally, with the puzzle pieces falling into place, Max discovered the true identity of the double agent. It was someone they had once trusted wholeheartedly, making the betrayal even more painful. Armed with this knowledge, Max contacted the agency's director, providing irrefutable evidence of the double agent's crimes. A secret operation was launched to apprehend the traitor, and Max played a pivotal role in bringing them to justice. The agency's internal security measures were strengthened, and Max's dedication to preserving global security was recognized by higher authorities. With the threat of the double agent neutralized, Max's attention turned back to The Phoenix. Undeterred by the previous setback, Max devised a new plan to gather intelligence on the organization. This time, Max would collaborate with international agencies and secret operatives from around the world. Together, they formed a united front against The Phoenix, vowing to dismantle the organization once and for all. Max Sterling's journey became a testament to resilience, resourcefulness, and the unyielding pursuit of justice. As Max dove deeper into the shadows, they became a beacon of hope for those whose lives were torn apart by the machinations of evil. The world waited with bated breath, knowing that their future hinged on Max's unwavering determination to eliminate the threat posed by The Phoenix.

\textbf{Chapter \#2:} Max Sterling, a skilled agent with a relentless determination, had made it his mission to bring down the notorious terrorist organization, "The Phoenix." Aware of a former member who had defected years ago, Max saw an opportunity to obtain valuable insider information that could cripple the organization once and for all. Determined to convince the defector to cooperate, Max embarked on a dangerous journey to track them down. Through countless hours of research, he pieced together clues and managed to locate the defector's current whereabouts. With a plan in mind, Max set out to approach the defector. He knew that his negotiating skills and persuasive tactics would be crucial in convincing them to share their knowledge. Max believed that with the right words, he could offer them a renewed sense of purpose and safety. As Max arrived at the defector's secluded hideout, a hint of nervous anticipation filled the air. He prepared himself mentally, going over the details of their previous affiliation with "The Phoenix" and the potential benefits of collaborating. Max believed that once the defector understood the impact their cooperation could have on dismantling the organization, they would surely be swayed. However, as Max began his carefully crafted speech, he quickly realized that it would be an uphill battle. The defector, haunted by their past experiences, was a broken soul. Their eyes reflected a deep-rooted fear, a visceral apprehension, that Max had not anticipated. They hesitated, their gaze filled with a mixture of fear and uncertainty. Despite Max's best efforts, the defector shook their head, unwilling to collaborate. They spoke of the constant threat "The Phoenix" posed to their own safety and that of their loved ones. Their voice quivered as they recalled the horrors they had witnessed under the organization's reign. To the defector, trusting anyone, even someone as skilled and determined as Max, seemed like an impossible feat. Max felt a twinge of disappointment, realizing the immense challenge he faced. He knew he couldn't force the defector to cooperate, and he respected their decision. It was clear that the wounds inflicted by "The Phoenix" ran deep, leaving scars that would not easily fade away. As he left the hideout, Max couldn't help but question his approach. He had hoped that his determination and persuasive techniques would be enough. Yet, he realized that sometimes, the damage caused by an organization's cruelty is too debilitating, leaving individuals unable to trust and risking their own lives and the lives of their loved ones in the process. But Max was not one to give up easily. He knew that he had to find another way to dismantle "The Phoenix" and protect innocent lives. He returned to his agency, determined to explore alternative approaches and gather intelligence from other sources. Max tirelessly worked alongside his team, chasing leads, infiltrating the organization, and gaining crucial information step by step. With their unwavering determination, they managed to uncover a considerable amount of evidence against "The Phoenix." As Max presented the evidence to higher-ups in the government, he found allies who shared his resolve. Together, they formed a multi-agency task force, pooling resources and skills to bring down the notorious organization once and for all. The collaboration between agencies was unprecedented, and it fostered an atmosphere of trust and cooperation that would prove vital in their mission. With the collective effort of the task force, they meticulously planned a high-risk operation to capture the key members of "The Phoenix" and gather enough evidence to put them behind bars for life. Max knew that it would require stealth, precision, and the element of surprise to succeed. Under the cover of darkness, the task force launched their operation. In a meticulously coordinated raid on multiple locations simultaneously, they apprehended the leaders of "The Phoenix" and seized valuable evidence that would ensure their downfall. News of the operation spread rapidly, and the world rejoiced at the dismantling of a terrorist organization that had plagued global security for years. Through the unwavering determination of Max Sterling and his dedicated colleagues, justice prevailed, and innocent lives were saved. Max Sterling's mission may not have gone as planned with the defector, but it taught him an important lesson about the power of trust, the debilitating effects of trauma, and the strength of collective effort. From that day forward, he understood that he couldn't do it alone. He needed the support of his team and the collaboration of others to tackle the darkest threats to global security and bring light to the world. Max Sterling became not just a skilled agent but a leader who understood the value of trust, unity, and determination.

\textbf{Chapter \#3:} Max Sterling had dedicated his life to hunting down the shadowy figure who led the international terrorist organization known as "The Phoenix." Their crimes had left a trail of devastation and despair around the world, and Max was determined to bring their reign of terror to an end. Frustration had gnawed at Max as his previous efforts had failed to yield any significant results. The Phoenix's encrypted communication channels remained impenetrable, shrouding their nefarious plans in secrecy. Max, however, possessed a unique skill set—a mastery of cryptography and hacking—that set him apart from the rest. Realizing that the clock was ticking and the world summit was drawing near, Max knew he had to take drastic action. In a dimly lit room filled with the soft hum of computers, he embarked on a relentless mission to breach The Phoenix's defenses. His fingers danced across the keyboard, his mind working in perfect harmony with the lines of code on the screen. It was a battle of wits as Max faced off against the organization's tech-savvy members. Layers of encryption were peeled back, firewalls breached, and digital traps skillfully avoided. Max's determination was unwavering. Days turned into nights as he delved deeper into the abyss of The Phoenix's digital world. The stakes were higher than ever, with the world summit fast approaching and the specter of a devastating attack looming. Failure was not an option. As Max progressed, he began to unravel The Phoenix's hidden communications. Messages, once obscured by complex algorithms, became clear. He identified potential weak links within the organization, individuals who could hold the key to unraveling the leader's identity and location. Each breakthrough brought him closer to his goal, but it also escalated the danger. The Phoenix, aware of Max's relentless pursuit, fought back with ferocity. They launched counterattacks, attempting to trace Max's location and expose him to their deadly network. But Max was relentless. He gathered a treasure trove of information about the organization's structure, their plans, and the identities of their key operatives. The evidence he collected was damning, and he knew it would be the key to dismantling The Phoenix once and for all. Finally, the moment of truth arrived. Max uncovered the location of The Phoenix's leader, a breakthrough that sent shivers down his spine. He had gathered enough evidence to not only bring down the organization but also to ensure the leader faced justice for their crimes. With the weight of the world on his shoulders, Max Sterling had succeeded in his mission. He had tracked down the elusive leader of The Phoenix and gathered the evidence needed to dismantle the organization. The world summit was safe, and innocent lives were spared from the impending disaster. As Max watched the authorities apprehend The Phoenix's leader, he knew that justice would prevail. The darkness that had shrouded The Phoenix was finally lifted, and a sense of closure washed over Max. His relentless pursuit of the truth had paid off, and he had achieved his mission to protect the world from the horrors of terrorism. In the end, Max Sterling's unwavering determination and mastery of technology had prevailed against the darkest of adversaries. The world owed a debt of gratitude to this enigmatic operative who had dedicated his life to ensuring peace and security prevailed over chaos and terror.

\end{document}